%% file: minilm.tex
\documentclass{article}
\usepackage{microtype}
\usepackage{graphicx}
\usepackage{subfigure}
\usepackage{booktabs} 

\usepackage[utf8]{inputenc} 
\usepackage[T1]{fontenc}    
\usepackage{url}            
\usepackage{booktabs}       
\usepackage{amsfonts}       
\usepackage{nicefrac}       

\usepackage{hyperref}

\usepackage[accepted]{icml2020}

\input{settings.tex}

\input{math_commands.tex}

\newcommand\ours{\textsc{MiniLM}}

\newcommand\bertbase{BERT$_\text{BASE}$}
\newcommand\bertlarge{BERT$_\text{LARGE}$}
\newcommand\xlmrbase{XLM-R$_\text{Base}$}
\newcommand{\robertabase}{RoBERTa$_{\textsc{base}}$}
\newcommand{\vonelarge}{\textsc{UniLM}$_{\textsc{large}}$}
\newcommand{\bartlarge}{BART$_{\textsc{large}}$}

\icmltitlerunning{\ours{}: Deep Self-Attention Distillation for Task-Agnostic Compression of Pre-Trained Transformers}

\begin{document}

\twocolumn[
\icmltitle{\ours{}: Deep Self-Attention Distillation for \\ Task-Agnostic Compression of Pre-Trained Transformers}




\begin{icmlauthorlist}
\icmlauthor{Wenhui Wang}{}
\icmlauthor{Furu Wei}{}
\icmlauthor{Li Dong}{}
\icmlauthor{Hangbo Bao}{}
\icmlauthor{Nan Yang}{}
\icmlauthor{Ming Zhou}{}
\end{icmlauthorlist}

\begin{icmlauthorlist}
Microsoft Research
\end{icmlauthorlist}

\begin{icmlauthorlist}
\tt \{wenwan,fuwei,lidong1,t-habao,nanya,mingzhou\}@microsoft.com
\end{icmlauthorlist}

\icmlcorrespondingauthor{Furu Wei}{fuwei@microsoft.com}
\icmlkeywords{minilm}
\vskip 0.3in
]



\printAffiliationsAndNotice{} 

\begin{abstract}

Pre-trained language models (e.g., BERT~\cite{bert} and its variants) have achieved remarkable success in varieties of NLP tasks. 
However, these models usually consist of hundreds of millions of parameters which brings challenges for fine-tuning and online serving in real-life applications 
due to latency and capacity constraints. 
In this work, we present a simple and effective approach to compress large Transformer~\cite{transformer} based pre-trained models, termed as deep self-attention distillation.
The small model (student) is trained by deeply mimicking the self-attention module, which plays a vital role in Transformer networks, of the large model (teacher). 
Specifically, we propose distilling the self-attention module of the last Transformer layer of the teacher, which is effective and flexible for the student. 
Furthermore, we introduce the scaled dot-product between values in the self-attention module as the new deep self-attention knowledge, in addition to the attention distributions (i.e., the scaled dot-product of queries and keys) that have been used in existing works.
Moreover, we show that introducing a teacher assistant~\cite{ta} also helps the distillation of large pre-trained Transformer models.
Experimental results demonstrate that our monolingual model\footnote{The code and models are publicly available at~\url{https://aka.ms/minilm}.} outperforms state-of-the-art baselines in different parameter size of student models. 
In particular, it retains more than $99\%$ accuracy on SQuAD 2.0 and several GLUE benchmark tasks using $50\%$ of the Transformer parameters and computations of the teacher model. 
We also obtain competitive results in applying deep self-attention distillation to multilingual pre-trained models.

\end{abstract}

\section{Introduction}

Language model (LM) pre-training has achieved remarkable success for various natural language processing tasks~\cite{elmo,ulmfit,gpt,bert,unilm,xlnet,spanbert,roberta}. The pre-trained language models, such as BERT~\cite{bert} and its variants, learn contextualized text representations by predicting words given their context using large scale text corpora, and can be fine-tuned with additional task-specific layers to adapt to downstream tasks. 
However, these models usually contain hundreds of millions of parameters which brings challenges for fine-tuning and online serving in real-life applications for latency and capacity constraints. 

Knowledge distillation~\cite{softlabeldistill,intermediatedistill} (KD) has been proven to be a promising way to compress a large model (called the teacher model) into a small model (called the student model), which uses much fewer parameters and computations while achieving competitive results on downstream tasks.
There have been some works that task-specifically distill pre-trained large LMs into small models~\cite{bert2lstm,studentInit,patientdistill,bertintermediate}. 
They first fine-tune the pre-trained LMs on specific tasks and then perform distillation. 
Task-specific distillation is effective, but fine-tuning large pre-trained models is still costly, especially for large datasets.
Different from task-specific distillation, 
task-agnostic LM distillation mimics the behavior of the original pre-trained LMs and the student model can be directly fine-tuned on downstream tasks~\cite{mminibert,distillbert,tinybert,mobilebert}.

Previous works use soft target probabilities for masked language modeling predictions or intermediate representations of the teacher LM to guide the training of the task-agnostic student.
DistilBERT~\cite{distillbert} employs a soft-label distillation loss and a cosine embedding loss, and initializes the student from the teacher by taking one layer out of two. But each Transformer layer of the student is required to have the same architecture as its teacher.    
TinyBERT~\cite{tinybert} and \textsc{MobileBERT}~\cite{mobilebert} utilize more fine-grained knowledge, including hidden states and self-attention distributions of Transformer networks, and transfer these knowledge to the student model layer-to-layer.
To perform layer-to-layer distillation, TinyBERT adopts a uniform function to determine the mapping between the teacher and student layers, and uses a parameter matrix to linearly transform student hidden states.
\textsc{MobileBERT} assumes the teacher and student have the same number of layers and introduces the bottleneck module to keep their hidden size the same. 

In this work, we propose the deep self-attention distillation framework for task-agnostic Transformer based LM distillation.
The key idea is to deeply mimic the self-attention modules which are the fundamentally important components in the Transformer based teacher and student models.
Specifically, we propose distilling the self-attention module of the last Transformer layer of the teacher model.
Compared with previous approaches, using knowledge of the last Transformer layer rather than performing layer-to-layer knowledge distillation alleviates the difficulties in layer mapping between the teacher and student models, and the layer number of our student model can be more flexible.
Furthermore, we introduce the scaled dot-product between values in the self-attention module as the new deep self-attention knowledge, in addition to the attention distributions (i.e., the scaled dot-product of queries and keys) that has been used in existing works. Using scaled dot-product between self-attention values also converts representations of different dimensions into relation matrices with the same dimensions  without introducing additional parameters to transform student representations,
allowing arbitrary hidden dimensions for the student model. 
Finally, we show that introducing a teacher assistant~\cite{ta} helps the distillation of large pre-trained Transformer based models and the proposed deep self-attention distillation can further boost the performance.

We conduct extensive experiments on downstream NLP tasks. Experimental results demonstrate that our monolingual model outperforms state-of-the-art baselines in different parameter size of student models. Specifically, the $6$-layer model of $768$ hidden dimensions distilled from \bertbase{} is $2.0\times$ faster, while retaining more than $99\%$ accuracy on SQuAD 2.0 and several GLUE benchmark tasks. Moreover, our multilingual model distilled from \xlmrbase{} also achieves competitive performance with much fewer Transformer parameters. 


\section{Preliminary}

Multi-layer Transformers~\cite{transformer} have been the most widely-used network structures in state-of-the-art pre-trained models.
In this section, we present a brief introduction to the Transformer network and the self-attention mechanism, which is the core component of the Transformer.
We also present the existing approaches on knowledge distillation for Transformer networks, particularly in the context of distilling a large Transformer based pre-trained model into a small Transformer model.

\subsection{Input Representation}
\label{sec:input}

Texts are tokenized to subword units by WordPiece~\cite{gnmt} in BERT~\cite{bert}. For example, the word \tx{forecasted} is split to \tx{forecast} and \tx{\#\#ed}, where \tx{\#\#} indicates the pieces are belong to one word.
A special boundary token \sptk{SEP} is used to separate segments if the input text contains more than one segment.
At the beginning of the sequence, a special token \sptk{CLS} is added to obtain the representation of the whole input.
The vector representations ($\{\textbf{x}_i\}_{i=1}^{|x|}$) of input tokens are computed via summing the corresponding token embedding, absolute position embedding, and segment embedding.

\subsection{Backbone Network: Transformer}
\label{sec:transformer}

Transformer~\cite{transformer} is used to encode contextual information for input tokens.
The input vectors $\{\textbf{x}_i\}_{i=1}^{|x|}$ are packed together into $\mathbf{H}^0 = [\mathbf{x}_1, \cdots, \mathbf{x}_{|x|}]$.
Then stacked Transformer blocks compute the encoding vectors as:
\begin{equation}
\mathbf{H}^l = \mathrm{Transformer}_{l}(\mathbf{H}^{l-1}),~l \in [1, L]
\end{equation}
where $L$ is the number of Transformer layers, and the final output is $\mathbf{H}^L = [\mathbf{h}_1^L, \cdots, \mathbf{h}_{|x|}^L]$.
The hidden vector $\mathbf{h}_i^L$ is used as the contextualized representation of $\mathbf{x}_i$.
Each Transformer layer consists of a self-attention sub-layer and a fully connected feed-forward network. Residual connection~\cite{resnet} is employed around each of the two sub-layers, followed by layer normalization~\cite{layernorm}.

\begin{figure*}[t]
\centering
\includegraphics[width=0.98\textwidth]{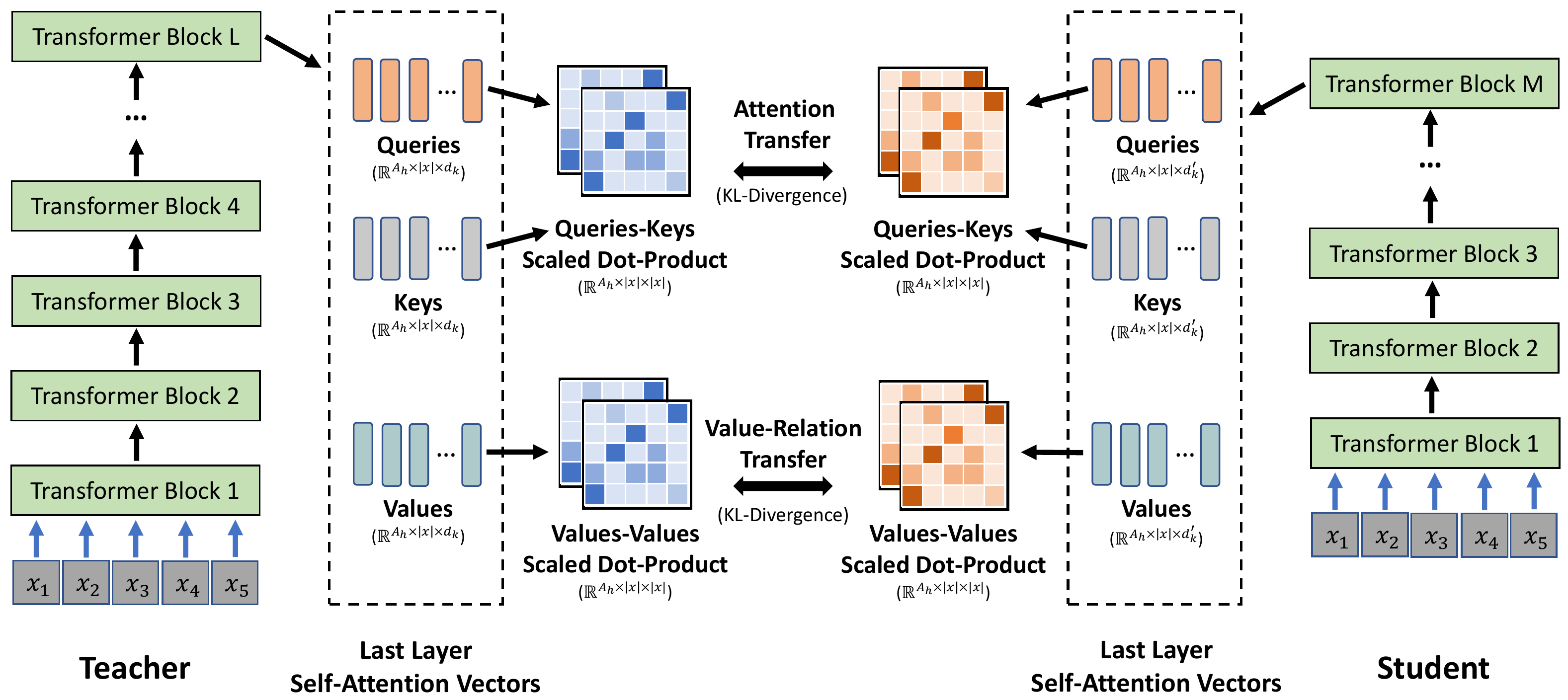}
\caption{Overview of Deep Self-Attention Distillation. The student is trained by deeply mimicking the self-attention behavior of the last Transformer layer of the teacher.
In addition to the self-attention distributions, we introduce the self-attention value-relation transfer to help the student achieve a deeper mimicry. Our student models are named as \ours{}.}
\label{fig:method}
\end{figure*}

\paragraph{Self-Attention}

In each layer, Transformer uses multiple self-attention heads to aggregate the output vectors of the previous layer.
For the $l$-th Transformer layer, the output of a self-attention head $\mathbf{AO}_{l,a},~a \in [1, A_h]$ is computed via:
\begin{gather}
\mathbf{Q}_{l,a} = \mathbf{H}^{l-1} \mathbf{W}_{l,a}^Q,~\mathbf{K}_{l,a} = \mathbf{H}^{l-1} \mathbf{W}_{l,a}^K,~\mathbf{V}_{l,a} = \mathbf{H}^{l-1} \mathbf{W}_{l,a}^V \\
\mathbf{A}_{l,a} = \softmax(\frac{\mathbf{Q}_{l,a} \mathbf{K}_{l,a}^{\intercal}}{ \sqrt{d_k}})  \\
\mathbf{AO}_{l,a} = \mathbf{A}_{l,a}\mathbf{V}_{l,a}
\end{gather}
where the previous layer's output $\mathbf{H}^{l-1} \in \mathbb{R}^{|x| \times d_h}$ is linearly projected to a triple of queries, keys and values using parameter matrices $\mathbf{W}_{l,a}^Q , \mathbf{W}_{l,a}^K , \mathbf{W}_{l,a}^V \in \mathbb{R}^{d_h \times d_k}$, respectively. $\mathbf{A}_{l,a} \in \mathbb{R}^{|x| \times |x|}$ indicates the attention distributions, which is computed by the scaled dot-product of queries and keys. $A_h$ represents the number of self-attention heads. $d_k \times A_h$ is equal to the hidden dimension $d_h$ in BERT.

\subsection{Transformer Distillation}

Knowledge distillation~\cite{softlabeldistill,intermediatedistill} is to train the small student model $S$ on a transfer feature set with soft labels and intermediate representations provided by the large teacher model $T$. Knowledge distillation is modeled as minimizing the differences between teacher and student features:
\begin{align}
\mathcal{L}_{\text{KD}} = {\sum_{e\in \mathcal{D}}{L(f^{S}(e), f^{T}(e))}} \label{eq:objective:kd}
\end{align}
Where $\mathcal{D}$ denotes the training data, $f^{S}(\cdot)$ and $f^{T}(\cdot)$ indicate the features of student and teacher models respectively, $L(\cdot)$ represents the loss function. 
The mean squared error (MSE) and KL-divergence are often used as loss functions.

For Transformer based LM distillation, soft target probabilities for masked language modeling predictions, embedding layer outputs, self-attention distributions and outputs (hidden states) of each Transformer layer of the teacher model are used as features to help the training of the student. Soft labels and embedding layer outputs are used in DistillBERT. 
TinyBERT and \textsc{MobileBERT} further utilize self-attention distributions and outputs of each Transformer layer.
For \textsc{MobileBERT}, the student is required to have the same number of layers as its teacher to perform layer-to-layer distillation. Besides, bottleneck and inverted bottleneck modules are introduced to keep the hidden size of the teacher and student are also the same. To transfer knowledge layer-to-layer, TinyBERT employs a uniform-function to map teacher and student layers.
Since the hidden size of the student can be smaller than its teacher, a parameter matrix is introduced to transform the student features.

\begin{table*}[t]
\caption{Comparison with previous task-agnostic Transformer based LM distillation approaches.}
\vskip 0.1in
\centering
\scalebox{0.85}{
\begin{tabular}{l c c c c c}
\toprule
\textbf{Approach}    & \textbf{Teacher Model}    & \textbf{Distilled Knowledge}    & \textbf{\tabincell{c}{Layer-to-Layer \\ Distillation}}   &  \textbf{\tabincell{c}{Requirements on \\ the number of \\ layers of students}}    & \textbf{\tabincell{c}{Requirements on \\ the hidden \\ size of students}} \\ \midrule
DistillBERT    & \bertbase{}    & \tabincell{c}{Soft target probabilities \\ Embedding outputs} &    &    & \checkmark \\ \cmidrule{3-3}
TinyBERT    & \bertbase{}    & \tabincell{c}{Embedding outputs \\ Hidden states \\ Self-Attention distributions}    & \checkmark   &    &  \\ \cmidrule{3-3}
\textsc{MobileBERT}    & \tabincell{c}{IB-\bertlarge{}}    &  \tabincell{c}{Soft target probabilities \\ Hidden states \\ Self-Attention distributions}    & \checkmark    & \checkmark    & \checkmark \\ \cmidrule{3-3}
\ours{}    & \bertbase{}    & \tabincell{c}{Self-Attention distributions \\ Self-Attention value relation}    &    &    &  \\ \bottomrule
\end{tabular}
}
\normalsize
\label{tbl:compare:distillation}
\vskip -0.1in
\end{table*}

\section{Deep Self-Attention Distillation}

Figure~\ref{fig:method} gives an overview of the deep self-attention distillation. 
The key idea is three-fold. First, we propose to train the student by deeply mimicking the self-attention module, which is the vital component in the Transformer, of the teacher's last layer.
Second, we introduce transferring the relation between values (i.e., the scaled dot-product between values) to achieve a deeper mimicry, in addition to performing attention distributions (i.e., the scaled dot-product of queries and keys) transfer in the self-attention module. 
Moreover, we show that introducing a teacher assistant~\cite{ta} also helps the distillation of large pre-trained Transformer models when the size gap between the teacher model and student model is large.

\subsection{Self-Attention Distribution Transfer}

The attention mechanism~\cite{attention} has been a highly successful neural network component for NLP tasks, which is also crucial for pre-trained LMs. Some works show that self-attention distributions of pre-trained LMs capture a rich hierarchy of linguistic information~\cite{bert_learns,bert_look_at}.
Transferring self-attention distributions has been used in previous works for Transformer distillation~\cite{tinybert,mobilebert,bertintermediate}. We also utilize the self-attention distributions to help the training of the student. Specifically, we minimize the KL-divergence between the self-attention distributions of the teacher and student:
\begin{align}
\mathcal{L}_{\text{AT}} = \frac{1}{A_h|x|}{\sum_{a=1}^{A_h}{\sum_{t=1}^{|x|}{D_{KL}(\mathbf{A}^{T}_{L,a,t} \parallel \mathbf{A}^{S}_{M,a,t})}}} \label{eq:objective:kd}
\end{align}
Where $|x|$ and $A_h$ represent the sequence length and the number of attention heads. $L$ and $M$ represent the number of layers for the teacher and student.
$\mathbf{A}^{T}_{L}$ and $\mathbf{A}^{S}_{M}$ are the attention distributions of the last Transformer layer for the teacher and student, respectively. They are computed by the scaled dot-product of queries and keys.

Different from previous works which transfer teacher's knowledge layer-to-layer, we only use the attention maps of the teacher's last Transformer layer. Distilling attention knowledge of the last Transformer layer allows more flexibility for the number of layers of our student models, avoids the effort of finding the best layer mapping. 

\subsection{Self-Attention Value-Relation Transfer}

In addition to the attention distributions, we propose using the relation between values in the self-attention module to guide the training of the student.
The value relation is computed via the multi-head scaled dot-product between values. The KL-divergence between the value relation of the teacher and student is used as the training objective:
\begin{gather}
\mathbf{VR}_{L,a}^T = \softmax(\frac{\mathbf{V}_{L,a}^{T} \mathbf{V}_{L,a}^{T\intercal}}{ \sqrt{d_k}})  \\
\mathbf{VR}_{M,a}^S = \softmax(\frac{\mathbf{V}_{M,a}^{S} \mathbf{V}_{M,a}^{S\intercal}}{ \sqrt{d_k^{\prime}}})  \\
\mathcal{L}_{\text{VR}} = \frac{1}{A_h|x|}{\sum_{a=1}^{A_h}{\sum_{t=1}^{|x|}{D_{KL}(\mathbf{VR}^{T}_{L,a,t} \parallel \mathbf{VR}^{S}_{M,a,t})}}} \label{eq:objective:kd}
\end{gather}
Where $\mathbf{V}_{L,a}^{T} \in \mathbb{R}^{|x| \times d_k}$ and $\mathbf{V}_{M,a}^{S} \in \mathbb{R}^{|x| \times d_k^{\prime}}$ are the values of an attention head in self-attention module for the teacher's and student's last Transformer layer. $\mathbf{VR}^{T}_{L} \in \mathbb{R}^{A_h \times |x| \times |x|}$ and $\mathbf{VR}^{S}_{M} \in \mathbb{R}^{A_h \times |x| \times |x|}$ are the value relation of the last Transformer layer for teacher and student, respectively.

The training loss is computed via summing the attention distribution transfer loss and value-relation transfer loss:
\begin{align}
\mathcal{L} = \mathcal{L}_{\text{AT}} + \mathcal{L}_{\text{VR}}
\end{align}

Introducing the relation between values enables the student to deeply mimic the teacher's self-attention behavior. Moreover, using the scaled dot-product converts vectors of different hidden dimensions into the relation matrices with the same size, which allows our students to use more flexible hidden dimensions and avoids introducing additional parameters to transform the student's representations.

\begin{table*}[t]
\caption{
Comparison between the publicly released $6$-layer models with 768 hidden size distilled from \bertbase{}.
We compare task-agnostic distilled models without task-specific distillation and data augmentation.
We report F1 for SQuAD 2.0, and accuracy for other datasets. The GLUE results of DistillBERT are taken from~\citet{distillbert}. We report the SQuAD 2.0 result by fine-tuning their released model$^{3}$.
For TinyBERT, we fine-tune the latest version of their public model$^{4}$ for a fair comparison. The results of our fine-tuning experiments are an average of $4$ runs for each task.
}
\vskip 0.1in
\begin{center}
\scalebox{0.93}{
\begin{tabular}{lcccccccccc}
\toprule
\bf Model & \bf \#Param & \bf SQuAD2 & \bf MNLI-m & \bf SST-2 & \bf QNLI & \bf CoLA & \bf RTE & \bf MRPC & \bf QQP & \bf Average \\ 
\midrule
\bertbase{} & 109M & 76.8 & 84.5 & 93.2 & 91.7 & 58.9 & 68.6 & 87.3 & 91.3 & 81.5 \\
DistillBERT & 66M & 70.7 & 79.0 & 90.7 & 85.3 & 43.6 & 59.9 & 87.5 & 84.9 & 75.2 \\
TinyBERT & 66M & 73.1 & 83.5 & 91.6 & 90.5 & 42.8 & \textbf{72.2} & \textbf{88.4} & 90.6 & 79.1 \\
\textbf{\textsc{MiniLM}} & 66M & \textbf{76.4}  & \textbf{84.0} & \textbf{92.0} & \textbf{91.0} & \textbf{49.2} & 71.5 & \textbf{88.4} & \textbf{91.0} & \textbf{80.4} \\
\bottomrule
\end{tabular}
}
\end{center}
\vskip -0.1in
\label{tbl:6l_768_exps}
\end{table*}

\subsection{Teacher Assistant}
Following~\citet{ta}, we introduce a teacher assistant (i.e., intermediate-size student model) to further improve the model performance of smaller students. 

Assuming the teacher model consists of $L$-layer Transformer with $d_{h}$ hidden size, the student model has $M$-layer Transformer with $d_{h}^{\prime}$ hidden size. 
For smaller students ($M\leq\frac{1}{2}L$, $d_{h}^{\prime}\leq\frac{1}{2}d_{h}$), we first distill the teacher into a teacher assistant with $L$-layer Transformer and $d_{h}^{\prime}$ hidden size. The assistant model is then used as the teacher to guide the training of the final student.
The introduction of a teacher assistant bridges the size gap between teacher and smaller student models, helps the distillation of Transformer based pre-trained LMs. 
Moreover, combining deep self-attention distillation with a teacher assistant brings further improvements for smaller student models.

\subsection{Comparison with Previous Work}

Table~\ref{tbl:compare:distillation} presents the comparison with previous approaches~\cite{distillbert,tinybert,mobilebert}. \textsc{MobileBERT} proposes using a specially designed inverted bottleneck model, which has the same model size as \bertlarge{}, as the teacher. The other methods utilize \bertbase{} to conduct experiments. For the knowledge used for distillation, our method introduces the scaled dot-product between values in the self-attention module as the new knowledge to deeply mimic teacher's self-attention behavior. TinyBERT and \textsc{MobileBERT} transfer knowledge of the teacher to the student layer-to-layer. \textsc{MobileBERT} assumes the student has the same number of layers as its teacher. TinyBERT employs a uniform strategy to determine its layer mapping. DistillBERT initializes the student with teacher's parameters, therefore selecting layers of the teacher model is still needed. \ours{} distills the self-attention knowledge of the teacher's last Transformer layer, which allows the flexible number of layers for the students and alleviates the effort of finding the best layer mapping. Student hidden size of DistillBERT and \textsc{MobileBERT} is required to be the same as its teacher. TinyBERT uses a parameter matrix to transform student hidden states. Using value relation allows our students to use arbitrary hidden size without introducing additional parameters.

\section{Experiments}

We conduct distillation experiments in different parameter size of student models, and evaluate the distilled models on downstream tasks including extractive question answering and the GLUE benchmark.

\begin{table*}[t]
\caption{
Comparison between student models of different architectures distilled from \bertbase{}. $M$ and $d_{h}^{\prime}$ indicate the number of layers and hidden dimension of the student model. TA indicates teacher assistant$^{5}$.
The fine-tuning results are averaged over $4$ runs.
}
\vskip 0.1in
\begin{center}
\scalebox{0.945}{
\begin{tabular}{cclcccc}
\toprule
\bf Architecture & \bf \#Param & \bf Model & \bf SQuAD 2.0 & \bf MNLI-m & \bf SST-2 & \bf Average \\ 
\midrule
\multirow{4}{*}{$M$=$6$;$d_{h}^{\prime}$=$384$} & \multirow{4}{*}{22M}
&MLM-KD (Soft-Label Distillation) & 67.9 & 79.6 & 89.8 & 79.1 \\
& &TinyBERT & 71.6 & 81.4 & 90.2 & 81.1 \\
& &\textsc{MiniLM} & 72.4  & 82.2 & 91.0 & 81.9 \\
& &\textsc{MiniLM} (w/ TA) & \textbf{72.7}  & \textbf{82.4} & \textbf{91.2} & \textbf{82.1} \\
\midrule
\multirow{4}{*}{$M$=$4$;$d_{h}^{\prime}$=$384$} & \multirow{4}{*}{19M}
&MLM-KD (Soft-Label Distillation) & 65.3 & 77.7 & 88.8 & 77.3 \\
& &TinyBERT & 66.7 & 79.2 & 88.5 & 78.1 \\
& &\textsc{MiniLM} & 69.4  & 80.3 & 90.2 & 80.0 \\
& &\textsc{MiniLM} (w/ TA) & \textbf{69.7}  & \textbf{80.6} & \textbf{90.6} & \textbf{80.3} \\
\midrule
\multirow{4}{*}{$M$=$3$;$d_{h}^{\prime}$=$384$} & \multirow{4}{*}{17M}
&MLM-KD (Soft-Label Distillation) & 59.9 & 75.2 & 88.0 & 74.4 \\
& &TinyBERT & 63.6 & 77.4 & 88.4 & 76.5 \\
& &\textsc{MiniLM} & 66.2  & 78.8 & 89.3 & 78.1 \\
& &\textsc{MiniLM} (w/ TA) & \textbf{66.9}  & \textbf{79.1} & \textbf{89.7} & \textbf{78.6} \\
\bottomrule
\end{tabular}
}
\end{center}
\vskip -0.1in
\label{tbl:diff_arcs}
\end{table*}

\begin{table}[t]
\caption{
The number of Embedding (Emd) and Transformer (Trm) parameters, and inference time for different models. 
}
\vskip 0.1in
\begin{center}
\scalebox{0.9}{
\begin{tabular}{ccccc}
\toprule
\bf \#Layers & \bf \tabincell{c}{Hidden \\ Size} & \bf \tabincell{c}{\#Param \\ (Emd)} & \bf \tabincell{c}{\#Param \\ (Trm)} & \bf \tabincell{c}{Inference \\ Time} \\ 
\midrule
12 & 768 & 23.4M & 85.1M & 93.1s (1.0$\times$) \\
6 & 768 & 23.4M & 42.5M & 46.9s (2.0$\times$) \\
12 & 384 & 11.7M & 21.3M & 34.8s (2.7$\times$) \\
6 & 384 & 11.7M & 10.6M & 17.7s (5.3$\times$) \\
4 & 384 & 11.7M & 7.1M & 12.0s (7.8$\times$) \\
3 & 384 & 11.7M & 5.3M & 9.2s (10.1$\times$) \\
\bottomrule
\end{tabular}
}
\end{center}
\vskip -0.1in
\label{tbl:infer_time}
\end{table}

\subsection{Distillation Setup}

We use the uncased version of \bertbase{} as our teacher.  BERT$_\text{BASE}$~\cite{bert} is a $12$-layer Transformer with $768$ hidden size, and $12$ attention heads, which contains about $109$M parameters. 
The number of heads of attention distributions and value relation are set to $12$ for student models.
We use documents of English Wikipedia\footnote{Wikipedia version: enwiki-20181101.} and BookCorpus~\cite{bookcorpus} for the pre-training data, following the preprocess and the WordPiece tokenization of~\citet{bert}. The vocabulary size is $30,522$. The maximum sequence length is $512$.
We use Adam~\cite{adam} with $\beta_1=0.9$, $\beta_2=0.999$. We train the $6$-layer student model with $768$ hidden size using $1024$ as the batch size and 5e-4 as the peak learning rate for $400,000$ steps. For student models of other architectures, the batch size and peak learning rate are set to $256$ and 3e-4, respectively. 
We use linear warmup over the first $4,000$ steps and linear decay.
The dropout rate is $0.1$. The weight decay is $0.01$. 

We also use an in-house pre-trained Transformer model in the \bertbase{} size as the teacher model, and distill it into $12$-layer and $6$-layer student models with $384$ hidden size. For the $12$-layer model, we use Adam~\cite{adam} with $\beta_1=0.9$, $\beta_2=0.98$. The model is trained using $2048$ as the batch size and 6e-4 as the peak learning rate for $400,000$ steps. The batch size and peak learning rate are set to $512$ and 4e-4 for the $6$-layer model. The rest hyper-parameters are the same as above BERT based distilled models.

For the training of multilingual \ours{} models, we use Adam~\cite{adam} with $\beta_1=0.9$, $\beta_2=0.999$. We train the $12$-layer student model using $256$ as the batch size and 3e-4 as the peak learning rate for $1,000,000$ steps. The $6$-layer student model is trained using $512$ as the batch size and 6e-4 as the peak learning rate for $400,000$ steps. 

We distill our student models using $8$ V100 GPUs with mixed precision training. Following~\citet{patientdistill} and~\citet{tinybert}, the inference time is evaluated on the QNLI training set with the same hyper-parameters. We report the average running time of $100$ batches on a single P100 GPU.

\subsection{Downstream Tasks}

Following previous language model pre-training~\cite{bert,roberta} and task-agnostic pre-trained language model distillation~\cite{distillbert,tinybert,mobilebert}, we evaluate our distilled models on the extractive question answering and GLUE benchmark.

\paragraph{Extractive Question Answering} Given a passage $P$, the task is to select a contiguous span of text in the passage by predicting its start and end positions to answer the question $Q$. We evaluate on SQuAD 2.0~\cite{squad2}, which has served as a major question answering benchmark.

Following BERT~\cite{bert}, we pack the question and passage tokens together with special tokens, to form the input: ``\sptk{CLS} $Q$ \sptk{SEP} $P$ \sptk{SEP}". Two linear output layers are introduced to predict the probability of each token being the start and end positions of the answer span. The questions that do not have an answer are treated as having an answer span with start and end at the \sptk{CLS} token.

\paragraph{GLUE} 

The General Language Understanding Evaluation (GLUE) benchmark~\cite{wang2018glue} consists of nine sentence-level classification tasks, including Corpus of Linguistic Acceptability (CoLA)~\cite{cola2018}, Stanford Sentiment Treebank (SST)~\cite{sst2013}, Microsoft Research Paraphrase Corpus
(MRPC)~\cite{mrpc2005}, Semantic Textual Similarity Benchmark (STS)~\cite{sts-b2017}, Quora Question Pairs (QQP)~\cite{chen2018quora}, Multi-Genre Natural Language Inference (MNLI)~\cite{mnli2017}, Question Natural Language Inference (QNLI)~\cite{squad1}, Recognizing Textual Entailment (RTE)~\cite{rte1,rte2,rte3,rte5} and Winograd Natural Language Inference (WNLI)~\cite{winograd2012}.
We add a linear classifier on top of the \sptk{CLS} token to predict label probabilities.

\subsection{Main Results}

Previous works~\cite{distillbert,patientdistill,tinybert} usually distill \bertbase{} into a $6$-layer student model with $768$ hidden size. We first conduct distillation experiments using the same student architecture. 
Results on SQuAD 2.0 and GLUE dev sets are presented in Table~\ref{tbl:6l_768_exps}.
Since \textsc{MobileBERT} distills a specially designed teacher with the inverted bottleneck modules, which has the same model size as \bertlarge{}, into a $24$-layer student using the bottleneck modules, we do not compare our models with \textsc{MobileBERT}.
\ours{} outperforms DistillBERT\footnote{The public model of DistillBERT is obtained from \url{https://github.com/huggingface/transformers/tree/master/examples/distillation}} and TinyBERT\footnote{We use the 2nd version TinyBERT from \url{https://github.com/huawei-noah/Pretrained-Language-Model/tree/master/TinyBERT}} across most tasks. Our model exceeds the two state-of-the-art models by $3.0$+$\%$ F1 on SQuAD 2.0 and $5.0$+$\%$ accuracy on CoLA.
We present the inference time for models in different parameter size in Table~\ref{tbl:infer_time}.
Our $6$-layer $768$-dimensional student model is $2.0\times$ faster than original \bertbase{}, while retaining more than $99\%$ performance on a variety of tasks, such as SQuAD 2.0 and MNLI.

We also conduct experiments for smaller student models. We compare \ours{} with our implemented MLM-KD (knowledge distillation using soft target probabilities for masked language modeling predictions) and TinyBERT, which are trained using the same data and hyper-parameters.
The results on SQuAD 2.0, MNLI and SST-2 dev sets are shown in Table~\ref{tbl:diff_arcs}. \ours{} outperforms soft label distillation and our implemented TinyBERT on the three tasks.
Deep self-attention distillation is also effective for smaller models.
Moreover, we show that introducing a teacher assistant\footnote{The teacher assistant is only introduced for the model \ours{} (w/ TA). The model \ours{} in different tables is directly distilled from its teacher model.} is also helpful in Transformer based pre-trained LM distillation, especially for smaller models. Combining deep self-attention distillation with a teacher assistant achieves further improvement for smaller student models.

\begin{table}[t]
\caption{
Effectiveness of self-attention value-relation (Value-Rel) transfer. The fine-tuning results are averaged over $4$ runs.
}
\vskip 0.1in
\begin{center}
\scalebox{0.85}{
\begin{tabular}{clccc}
\toprule
\bf Architecture & \bf Model & \bf SQuAD2 & \bf MNLI-m & \bf SST-2 \\ 
\midrule
\multirow{2}{*}{$M$=$6$;$d_{h}^{\prime}$=$384$}
&\textsc{MiniLM} & \textbf{72.4}  & \textbf{82.2} & \textbf{91.0} \\
&~~-Value-Rel & 71.0 & 80.9 & 89.9 \\
\midrule
\multirow{2}{*}{$M$=$4$;$d_{h}^{\prime}$=$384$}
&\textsc{MiniLM} & \textbf{69.4}  & \textbf{80.3} & \textbf{90.2} \\
&~~-Value-Rel & 67.5 & 79.0 & 89.2 \\
\midrule
\multirow{2}{*}{$M$=$3$;$d_{h}^{\prime}$=$384$}
&\textsc{MiniLM} & \textbf{66.2}  & \textbf{78.8} & \textbf{89.3} \\
&~~-Value-Rel & 64.2  & 77.8 & 88.3 \\
\bottomrule
\end{tabular}
}
\end{center}
\vskip -0.1in
\label{tbl:ablation}
\end{table}

\subsection{Ablation Studies}

We do ablation tests on several tasks to analyze the contribution of self-attention value-relation transfer. The dev results of SQuAD 2.0, MNLI and SST-2 are illustrated in Table~\ref{tbl:ablation}, using self-attention value-relation transfer positively contributes to the final results for student models in different parameter size. Distilling the fine-grained knowledge of value relation helps the student model deeply mimic the self-attention behavior of the teacher, which further improves model performance.  

We also compare different loss functions over values in the self-attention module. We compare our proposed value relation with mean squared error (MSE) over the teacher and student values. An additional parameter matrix is introduced to transform student values if the hidden dimension of the student is smaller than its teacher. The dev results on three tasks are presented in Table~\ref{tbl:ablation_mse}. Using value relation achieves better performance. Specifically, our method brings about $1.0\%$ F1 improvement on the SQuAD benchmark.
Moreover, there is no need to introduce additional parameters for our method. 
We have also tried to transfer the relation between hidden states. But we find the performance of student models are unstable for different teacher models. 

To show the effectiveness of distilling self-attention knowledge of the teacher's last Transformer layer, we compare our method with layer-to-layer distillation. We transfer the same knowledge and adopt a uniform strategy as in~\citet{tinybert} to map teacher and student layers to perform layer-to-layer distillation. The dev results on three tasks are presented in Table~\ref{tbl:ablation_ltl}.
\ours{} achieves better results.
It also alleviates the difficulties in layer mapping between the teacher and student. Besides, distilling the teacher's last Transformer layer requires less computation than layer-to-layer distillation, results in faster training speed.

\begin{table}[t]
\caption{
Comparison between different loss functions: KL-divergence over the value relation (the scaled dot-product between values) and mean squared error (MSE) over values. A parameter matrix is introduced to transform student values to have the same dimensions as the teacher values~\cite{tinybert}. The fine-tuning results are an average of $4$ runs for each task.
}
\vskip 0.1in
\begin{center}
\scalebox{0.85}{
\begin{tabular}{clccc}
\toprule
\bf Architecture & \bf Model & \bf SQuAD2 & \bf MNLI-m & \bf SST-2 \\ 
\midrule
\multirow{2}{*}{$M$=$6$;$d_{h}^{\prime}$=$384$}
&\textsc{MiniLM} & \textbf{72.4}  & \textbf{82.2} & \textbf{91.0} \\
&Value-MSE & 71.4 & 82.0 & 90.8 \\
\midrule
\multirow{2}{*}{$M$=$4$;$d_{h}^{\prime}$=$384$}
&\textsc{MiniLM} & \textbf{69.4}  & \textbf{80.3} & \textbf{90.2} \\
&Value-MSE & 68.3 & 80.1 & 89.9 \\
\midrule
\multirow{2}{*}{$M$=$3$;$d_{h}^{\prime}$=$384$}
&\textsc{MiniLM} & \textbf{66.2}  & \textbf{78.8} & \textbf{89.3} \\
&Value-MSE & 65.5  & 78.4 & \textbf{89.3} \\
\bottomrule
\end{tabular}
}
\end{center}
\vskip -0.1in
\label{tbl:ablation_mse}
\end{table}

\begin{table*}[t]
\caption{
Comparison between distilling knowledge of the teacher's last Transformer layer and layer-to-layer distillation. We adopt a uniform strategy as in~\citet{tinybert} to determine the mapping between teacher and student layers. The fine-tuning results are an average of $4$ runs for each task.
}
\vskip 0.1in
\begin{center}
\scalebox{0.945}{
\begin{tabular}{clcccc}
\toprule
\bf Architecture & \bf Model & \bf SQuAD 2.0 & \bf MNLI-m & \bf SST-2 & \bf Average \\ 
\midrule
\multirow{2}{*}{$M$=$6$;$d_{h}^{\prime}$=$384$}
&\textsc{MiniLM} & \textbf{72.4}  & \textbf{82.2} & \textbf{91.0} & \textbf{81.9} \\
&~~+Layer-to-Layer Distillation & 71.6 & 81.8 & 90.6 & 81.3 \\
\midrule
\multirow{2}{*}{$M$=$4$;$d_{h}^{\prime}$=$384$}
&\textsc{MiniLM} & \textbf{69.4}  & \textbf{80.3} & \textbf{90.2} & \textbf{80.0} \\
&~~+Layer-to-Layer Distillation & 67.6 & 79.9 & 89.6 & 79.0 \\
\midrule
\multirow{2}{*}{$M$=$3$;$d_{h}^{\prime}$=$384$}
&\textsc{MiniLM} & \textbf{66.2}  & \textbf{78.8} & \textbf{89.3} & \textbf{78.1} \\
&~~+Layer-to-Layer Distillation & 64.8  & 77.7 & 88.6 & 77.0 \\
\bottomrule
\end{tabular}
}
\end{center}
\label{tbl:ablation_ltl}
\vskip -0.1in
\end{table*}

\begin{table*}[t]
\caption{
The results of \ours{} distilled from an in-house pre-trained Transformer model (\bertbase{} size, $12$-layer Transformer, $768$-hidden size, and $12$ self-attention heads) on SQuAD 2.0 and GLUE benchmark. We report our $12$-layer$^{a}$ and $6$-layer$^{b}$ models with $384$ hidden size. The fine-tuning results are averaged over $4$ runs.
}
\vskip 0.1in
\begin{center}
\scalebox{0.93}{
\begin{tabular}{lcccccccccc}
\toprule
\bf Model & \bf \#Param & \bf SQuAD2 & \bf MNLI-m & \bf SST-2 & \bf QNLI & \bf CoLA & \bf RTE & \bf MRPC & \bf QQP & \bf Average \\ 
\midrule
\bertbase{} & 109M & 76.8 & 84.5 & 93.2 & 91.7 & 58.9 & 68.6 & 87.3 & 91.3 & 81.5 \\
\ours{}$^{a}$ & 33M & 81.7  & 85.7 & 93.0 & 91.5 & 58.5 & 73.3 & 89.5 & 91.3 & 83.1 \\
\ours{}$^{b}$ (w/ TA) & 22M & 75.6  & 83.3 & 91.5 & 90.5 & 47.5 & 68.8 & 88.9 & 90.6 & 79.6 \\
\bottomrule
\end{tabular}
}
\end{center}
\vskip -0.1in
\label{tbl:ulmv2_exps}
\end{table*}






\begin{table}[t]
\caption{
Question generation results of our $12$-layer$^{a}$ and $6$-layer$^{b}$ models with $384$ hidden size on SQuAD 1.1.
The first block follows the data split in~\citet{du-qg-2018}, while the second block is the same as in~\citet{zhao-qg-2018}.
MTR is short for METEOR, RG for ROUGE, and B for BLEU. 
}
\vskip 0.1in
\centering
\scalebox{0.88}{
\begin{tabular}{l c c c c}
\toprule
                   &  \textbf{\#Param} & \textbf{B-4} & \textbf{MTR}   & \textbf{RG-L}  \\ \midrule
\cite{du-qg-2018}  &  & 15.16           & 19.12          & -              \\
\cite{zhang-qg-2019} &  & 18.37           & 22.65          & 46.68          \\
\vonelarge{}        & 340M  & 22.78           & 25.49          & 51.57          \\
\ours{}$^{a}$       & 33M   & 21.07 & 24.09 & 49.14 \\
\ours{}$^{b}$ (w/ TA)     & 22M     & 20.31 & 23.43 & 48.21 \\
\midrule
\cite{zhao-qg-2018}  &  & 16.38           & 20.25          & 44.48          \\
\cite{zhang-qg-2019}  &  & 20.76           & 24.20          & 48.91          \\
\vonelarge{}       & 340M   & 24.32           & 26.10          & 52.69          \\
\ours{}$^{a}$        & 33M  & 23.27 & 25.15 & 50.60 \\
\ours{}$^{b}$ (w/ TA)      & 22M     & 22.01 & 24.24 & 49.51 \\ \bottomrule
\end{tabular}
}
\vskip -0.1in
\label{tbl:qg_exps}
\end{table}




\begin{table*}[t]
\caption{
Abstractive summarization results of our $12$-layer$^{a}$ and $6$-layer$^{b}$ models with $384$ hidden size on CNN/DailyMail and XSum.
The evaluation metric is the F1 version of ROUGE (RG) scores.
}
\vskip 0.1in
\centering
\begin{tabular}{lclccclccc}
\toprule
\multirow{2}{*}{\bf Model}                          & \multirow{2}{*}{\bf \#Param} & & \multicolumn{3}{c}{\bf CNN/DailyMail} & & \multicolumn{3}{c}{\bf XSum} \\
&                              & & RG-1  & RG-2  &      RG-L      & & RG-1  & RG-2  &     RG-L     \\ \midrule
\textsc{Lead}-3 &                                                         & & 40.42 & 17.62 &     36.67      & & 16.30 & 1.60  &    11.95     \\
\textsc{PtrNet}~\cite{see-2017-get} &                                      & & 39.53 & 17.28 &     36.38      & & 28.10 & 8.02  &    21.72     \\
Bottom-Up~\cite{gehrmann-etal-2018-bottom} &                                      & & 41.22 & 18.68 &     38.34      & & - & -  &    -     \\
 \vonelarge{}~\cite{unilm}                                        & 340M                         & & 43.08 & 20.43 &     40.34      & &   -   &   -   &      -       \\
 \bartlarge{}~\cite{bart}                                        & 400M                         & & 44.16 & 21.28 &     40.90      & & 45.14 & 22.27 &    37.25     \\
 {T5$_{\textsc{11B}}$}~\cite{t5}                               & 11B                          & & 43.52 & 21.55 & 40.69 & &   -   &   -   &      -       \\ 
{MASS$_{\textsc{base}}$}~\cite{mass}                            & 123M                         & & 42.12 & 19.50 &     39.01      & & 39.75 & 17.24 &    31.95     \\
\textsc{BERTSumAbs}~\cite{bertsum}                                 & 156M                         & & 41.72 & 19.39 &     38.76      & & 38.76 & 16.33 &    31.15     \\
{T5$_{\textsc{base}}$}~\cite{t5}                              & 220M                         & &   42.05 & 20.34 & 39.40        & &   -   &   -   &      -       \\ \midrule
\ours{}$^{a}$                                          & 33M                         & &   42.66 & 19.91 & 39.73 & & 40.43 & 17.72 &    32.60     \\
\ours{}$^{b}$ (w/ TA)                                          & 22M                         & &   41.57 & 19.21 & 38.64 & & 38.79 & 16.39 &    31.10     \\
\bottomrule
\end{tabular}
\vskip -0.1in
\label{tbl:summ_exps}
\end{table*}

\begin{table*}[t]
\caption{
Cross-lingual classification results of our $12$-layer$^{a}$ and $6$-layer$^{b}$ multilingual models with $384$ hidden size on XNLI. We report the accuracy on each of the 15 XNLI languages and the average accuracy. 
Results of mBERT, XLM-100 and \xlmrbase{} are from~\citet{xlmr}. 
}
\vskip 0.1in
\begin{center}
\scalebox{0.73}{
\begin{tabular}{l|cc|ccccccccccccccc|c}
\toprule
\bf Model & \bf \#Layers & \bf \#Hidden & \bf en & \bf fr & \bf es & \bf de & \bf el & \bf bg & \bf ru & \bf tr & \bf ar & \bf vi & \bf th & \bf zh & \bf hi & \bf sw & \bf ur & \bf Avg \\ 
\midrule
mBERT & 12 & 768 & 82.1 & 73.8 & 74.3 & 71.1 & 66.4 & 68.9 & 69.0 & 61.6 & 64.9 & 69.5 & 55.8 & 69.3 & 60.0 & 50.4 & 58.0 & 66.3 \\
XLM-$100$ & 16 & 1280 & 83.2 & 76.7 & 77.7 & 74.0 & 72.7 & 74.1 & 72.7 & 68.7 & 68.6 & 72.9 & 68.9 & 72.5 & 65.6 & 58.2 & 62.4 & 70.7 \\
\xlmrbase{} & 12 & 768 & 84.6  & 78.4 & 78.9 & 76.8 & 75.9 & 77.3 & 75.4 & 73.2 & 71.5 & 75.4 & 72.5 & 74.9 & 71.1 & 65.2 & 66.5 & 74.5 \\
\ours{}$^{a}$ & 12 & 384 & 81.5  & 74.8 & 75.7 & 72.9 & 73.0 & 74.5 & 71.3 & 69.7 & 68.8 & 72.1 & 67.8 & 70.0 & 66.2 & 63.3 & 64.2 & 71.1 \\
\ours{}$^{b}$ (w/ TA) & 6 & 384 & 79.2  & 72.3 & 73.1 & 70.3 & 69.1 & 72.0 & 69.1 & 64.5 & 64.9 & 69.0 & 66.0 & 67.8 & 62.9 & 59.0 & 60.6 & 68.0 \\
\bottomrule
\end{tabular}
}
\end{center}
\vskip -0.1in
\label{tbl:xnli_exps}
\end{table*}

\begin{table}[t]
\caption{
The number of Transformer (Trm) and Embedding (Emd) parameters for different multilingual pre-trained models and our distilled models.
}
\vskip 0.1in
\begin{center}
\scalebox{0.8}{
\begin{tabular}{lccccc}
\toprule
\bf Model & \bf \#Layers & \bf \tabincell{c}{Hidden \\ Size} & \bf \#Vocab & \bf \tabincell{c}{\#Param \\ (Trm)} & \bf \tabincell{c}{\#Param \\ (Emd)} \\ 
\midrule
mBERT & 12 & 768 & 110k & 85M & 85M \\
XLM-15 & 12 & 1024 & 95k & 151M & 97M \\
XLM-100 & 16 & 1280 & 200k & 315M & 256M \\
\xlmrbase{} & 12 & 768 & 250k & 85M & 192M \\
\ours{}$^{a}$ & 12 & 384 & 250k & 21M & 96M \\
\ours{}$^{b}$ & 6 & 384 & 250k & 11M & 96M \\

\bottomrule
\end{tabular}
}
\end{center}
\vskip -0.1in
\label{tbl:multilingual_models_params}
\end{table}

\begin{table*}[t]
\caption{
Cross-lingual question answering results of our $12$-layer$^{a}$ and $6$-layer$^{b}$ multilingual models with $384$ hidden size on MLQA. We report the F1 and EM (exact match) scores on each of the 7 MLQA languages. Results of mBERT and XLM-$15$ are taken from~\citet{mlqa}. $\dag$ indicates results of \xlmrbase{} taken from~\citet{xlmr}. 
We also report our fine-tuned results ($\ddag$) of \xlmrbase{}. 
}
\vskip 0.1in
\begin{center}
\scalebox{0.75}{
\begin{tabular}{l|cc|ccccccc|c}
\toprule
\bf Model & \bf \#Layers & \bf \#Hidden & \bf en & \bf es & \bf de & \bf ar & \bf hi & \bf vi & \bf zh & \bf Avg \\ 
\midrule
mBERT & 12 & 768 & 77.7 / 65.2 & 64.3 / 46.6 &  57.9 / 44.3 & 45.7 / 29.8 &  43.8 / 29.7 & 57.1 / 38.6 & 57.5 / 37.3 & 57.7 / 41.6 \\
XLM-$15$ & 12 & 1024 & 74.9 / 62.4 & 68.0 / 49.8 & 62.2 / 47.6 & 54.8 / 36.3 & 48.8 / 27.3 &  61.4 / 41.8 & 61.1 / 39.6 & 61.6 / 43.5 \\
\xlmrbase{}$\dag$ & 12 & 768 & 77.8 / 65.3  & 67.2 / 49.7 & 60.8 / 47.1 & 53.0 / 34.7 & 57.9 / 41.7 & 63.1 / 43.1 & 60.2 / 38.0 & 62.9 / 45.7 \\
\xlmrbase{}$\ddag$ & 12 & 768 & 80.3 / 67.4  & 67.0 / 49.2 & 62.7 / 48.3 & 55.0 / 35.6 & 60.4 / 43.7 & 66.5 / 45.9 & 62.3 / 38.3 & 64.9 / 46.9 \\
\ours{}$^{a}$ & 12 & 384 & 79.4 / 66.5 & 66.1 / 47.5 &  61.2 / 46.5 & 54.9 / 34.9 &  58.5 / 41.3 & 63.1 / 42.1 & 59.0 / 33.8 & 63.2 / 44.7 \\
\ours{}$^{b}$ (w/ TA) & 6 & 384 & 75.5 / 61.9 & 55.6 / 38.2 &  53.3 / 37.7 & 43.5 / 26.2 &  46.9 / 31.5 & 52.0 / 33.1 & 48.8 / 27.3 & 53.7 / 36.6 \\
\bottomrule
\end{tabular}
}
\end{center}
\vskip -0.1in
\label{tbl:mlqa_exps}
\end{table*}

\section{Discussion}

\subsection{Better Teacher Better Student}
We report the results of \ours{} distilled from an in-house pre-trained Transformer model following \textsc{UniLM}~\cite{unilm,unilmv2} in the \bertbase{} size. The teacher model is trained using similar pre-training datasets as in \robertabase{}~\cite{roberta}, which includes $160$GB text corpora from English Wikipedia, BookCorpus~\cite{bookcorpus}, OpenWebText\footnote{\url{skylion007.github.io/OpenWebTextCorpus}}, CC-News~\cite{roberta}, and Stories~\cite{stories_data}. We distill the teacher model into $12$-layer and $6$-layer models with $384$ hidden size using the same corpora. The $12$x$384$ model is used as the teacher assistant to train the 
$6$x$384$ model. We present the dev results of SQuAD 2.0 and GLUE benchmark in Table~\ref{tbl:ulmv2_exps}, the results of \ours{} are significantly improved. The $12$x$384$ \ours{} achieves $2.7\times$ speedup while performs competitively better than \bertbase{} in SQuAD 2.0 and GLUE benchmark datasets. 

\subsection{\ours{} for NLG Tasks}
We also evaluate \ours{} on natural language generation tasks, such as question generation and abstractive summarization. Following~\citet{unilm}, we fine-tune \ours{} as a sequence-to-sequence model by employing a specific self-attention mask. 

\paragraph{Question Generation} We conduct experiments for the answer-aware question generation task~\cite{du-qg-2018}. Given an input passage and an answer, the task is to generate a question that asks for the answer. The SQuAD 1.1 dataset~\cite{squad1} is used for evaluation. The results of \ours{}, \vonelarge{} and several state-of-the-art models are presented in Table~\ref{tbl:qg_exps}, our $12$x$384$ and $6$x$384$ distilled models achieve competitive performance on the question generation task.

\paragraph{Abstractive Summarization} We evaluate \ours{} on two abstractive summarization datasets, i.e., XSum~\cite{xsum}, and the non-anonymized version of CNN/DailyMail~\cite{see-2017-get}. The generation task is to condense a document into a concise and fluent summary, while conveying its key information. We report ROUGE scores~\cite{lin-2004-rouge} on the datasets.
Table~\ref{tbl:summ_exps} presents the results of \ours{}, baseline, several state-of-the-art models and pre-trained Transformer models. Our $12$x$384$ model outperforms BERT based method \textsc{BERTSumAbs}~\cite{bertsum} and the pre-trained sequence-to-sequence model {MASS$_{\textsc{base}}$}~\cite{mass} with much fewer parameters. Moreover, our $6$x$384$ \ours{} also achieves competitive performance.

\subsection{Multilingual \ours{}}

We conduct experiments on task-agnostic knowledge distillation of multilingual pre-trained models. 
We use the \xlmrbase{}\footnote{We use the v$0$ version of \xlmrbase{} in our distillation and fine-tuning experiments.}~\cite{xlmr} as the teacher and distill the model into $12$-layer and $6$-layer models with 384 hidden size using the same corpora. 
The $6$x$384$ model is trained using the $12$x$384$ model as the teacher assistant.
Given the vocabulary size of multilingual pre-trained models is much larger than monolingual models (30k for monolingual BERT, 250k for XLM-R), soft-label distillation for multilingual pre-trained models requires more computation. \ours{} only uses the deep self-attention knowledge of the teacher's last Transformer layer. The training speed of \ours{} is much faster than soft-label distillation for multilingual pre-trained models. 

We evaluate the student models on cross-lingual natural language inference (XNLI) benchmark~\cite{xnli} and cross-lingual question answering (MLQA) benchmark~\cite{mlqa}.

\paragraph{XNLI} Table~\ref{tbl:xnli_exps} presents XNLI results of our distilled students and several pre-trained LMs. 
Following~\citet{xlmr}, we select the best single model on
the joint dev set of all the languages. 
We present the number of Transformer and embedding parameters for different multilingual pre-trained models and our distilled models in Table~\ref{tbl:multilingual_models_params}.
\ours{} achieves competitive performance on XNLI with much fewer Transformer parameters. Moreover, the $12$x$384$ \ours{} compares favorably with mBERT~\cite{bert} and XLM~\cite{xlm} trained on the MLM objective. 

\paragraph{MLQA} Table~\ref{tbl:mlqa_exps} shows cross-lingual question answering results.
Following~\citet{mlqa}, we adopt SQuAD 1.1 as training data and use MLQA English development data for early stopping. The $12$x$384$ \ours{} performs competitively better than mBERT and XLM. Our $6$-layer \ours{} also achieves competitive performance. 

\section{Related Work}

\subsection{Pre-trained Language Models}

Unsupervised pre-training of language models~\cite{elmo,ulmfit,gpt,bert,clozepretrain19,mass,unilm,xlnet,spanbert,roberta,bart,t5} has achieved significant improvements for a wide range of NLP tasks. Early methods for pre-training~\cite{elmo,gpt} were based on standard language models. 
Recently, BERT~\cite{bert} proposes to use a masked language modeling objective to train a deep bidirectional Transformer encoder, which learns interactions between left and right context. 
\citet{roberta} show that very strong performance can be achieved by training the model longer over more data. 
\citet{spanbert} extend BERT by masking contiguous random spans. 
\citet{xlnet} predict masked tokens auto-regressively in a permuted order. 

To extend the applicability of pre-trained Transformers for NLG tasks. 
\citet{unilm} extend BERT by utilizing specific
self-attention masks to jointly optimize bidirectional, unidirectional and sequence-to-sequence masked language modeling objectives.
\citet{t5} employ an encoder-decoder Transformer and perform sequence-to-sequence pre-training by predicting the masked tokens in the encoder and decoder. 
Different from~\citet{t5}, \citet{bart} predict tokens auto-regressively in the decoder. 

\subsection{Knowledge Distillation}

Knowledge distillation has proven a promising way to compress large models while maintaining accuracy. It transfers the knowledge of a large model or an ensemble of neural networks (teacher) to a single lightweight model (student). \citet{softlabeldistill} first propose transferring the knowledge of the teacher to the student by using its soft target distributions to train the distilled model. \citet{intermediatedistill} introduce intermediate representations from hidden layers of the teacher to guide the training of the student. Knowledge of the attention maps~\cite{att_dst,att_dst_mrc} is also introduced to help the training.

In this work, we focus on task-agnostic knowledge distillation of large pre-trained Transformer based language models. There have been some works that task-specifically distill the fine-tuned language models on downstream tasks. \citet{bert2lstm} distill fine-tuned BERT into an extremely small bidirectional LSTM. \citet{dst_lm_init} initialize the student with a small pre-trained LM during task-specific distillation. \citet{patientdistill} introduce the hidden states from every $k$ layers of the teacher to perform knowledge distillation layer-to-layer. \citet{bertintermediate} further introduce the knowledge of self-attention distributions and propose progressive and stacked distillation methods.
Task-specific distillation requires to first fine-tune the large pre-trained LMs on downstream tasks and then perform knowledge transfer. The procedure of fine-tuning large pre-trained LMs is costly and time-consuming, especially for large datasets. 

For task-agnostic distillation, the distilled model mimics the original large pre-trained LM and can be directly fine-tuned on downstream tasks. In practice, task-agnostic compression of pre-trained LMs is more desirable. 
MiniBERT~\cite{mminibert} uses the soft target distributions for masked language modeling predictions to guide the training of the multilingual student model and shows its effectiveness on sequence labeling tasks.
DistillBERT~\cite{distillbert} uses the soft label and embedding outputs of the teacher to train the student.
TinyBERT~\cite{tinybert} and \textsc{MobileBERT}~\cite{mobilebert} further introduce self-attention distributions and hidden states to train the student. \textsc{MobileBERT} employs inverted bottleneck and bottleneck modules for teacher and student to make their hidden dimensions the same.   
The student model of \textsc{MobileBERT} is required to have the same number of layers as its teacher to perform layer-to-layer distillation. Besides, \textsc{MobileBERT} proposes a bottom-to-top progressive scheme to transfer teacher's knowledge.
TinyBERT uses a uniform-strategy to map the layers of teacher and student when they have different number of layers, and a linear matrix is introduced to transform the student hidden states to have the same dimensions as the teacher. TinyBERT also introduces task-specific distillation and data augmentation for downstream tasks, which brings further improvements. 

Different from previous works, our method employs the self-attention distributions and value relation of the teacher's last Transformer layer to help the student deeply mimic the self-attention behavior of the teacher. Using knowledge of the last Transformer layer instead of layer-to-layer distillation avoids restrictions on the number of student layers and the effort of finding the best layer mapping. 
Distilling relation between self-attention values allows the hidden size of students to be more flexible and avoids introducing linear matrices to transform student representations. 

\section{Conclusion}

In this work, we propose a simple and effective knowledge distillation method to compress large pre-trained Transformer based language models. The student is trained by deeply mimicking the teacher's self-attention modules, which are the vital components of the Transformer networks. We propose using the self-attention distributions and value relation of the teacher's last Transformer layer to guide the training of the student
, which is effective and flexible for the student models.
Moreover, we show that introducing a teacher assistant also helps pre-trained Transformer based LM distillation, and the proposed deep self-attention distillation can further boost the performance.
Our student model distilled from \bertbase{} retains high accuracy on SQuAD 2.0 and the GLUE benchmark tasks, and outperforms state-of-the-art baselines. The deep self-attention distillation can also be applied to compress pre-trained models in larger size. We leave it as our future work.

\bibliography{minilm}
\bibliographystyle{icml2020}

\appendix

\section{GLUE Benchmark}

The summary of datasets used for the General Language Understanding Evaluation (GLUE) benchmark\footnote{\url{https://gluebenchmark.com/}}~\cite{wang2018glue} is presented in Table~\ref{tbl:glue:datasets}.

We present the dataset statistics and metrics of SQuAD 2.0\footnote{\url{http://stanford-qa.com}}~\cite{squad2} in Table~\ref{tbl:squad2}.

\section{Fine-tuning Hyper-parameters}

\paragraph{Extractive Question Answering} For SQuAD 2.0, the maximum sequence length is $384$ and a sliding window of size $128$ if the lengths are longer than $384$. For the $12$-layer model distilled from our in-house pre-trained model, we fine-tune $3$ epochs using $48$ as the batch size and 4e-5 as the peak learning rate. The rest distilled models are trained using $32$ as the batch size and 6e-5 as the peak learning rate for $3$ epochs. 

\paragraph{GLUE} The maximum sequence length is $128$ for the GLUE benchmark. We set batch size to $32$, choose learning rates from \{2e-5, 3e-5, 4e-5, 5e-5\} and epochs from \{$3$, $4$, $5$\} for student models distilled from \bertbase{}. For student models distilled from our in-house pre-trained model, the batch size is chosen from \{$32$, $48$\}. We fine-tune several tasks (CoLA, RTE and MRPC) with longer epochs (up to $10$ epochs), which brings slight improvements. For the $12$-layer model, the learning rate used for CoLA, RTE and MRPC tasks is 1.5e-5. 

\begin{table}[t]
\caption{Summary of the GLUE benchmark.
}
\vskip 0.1in
\centering
\small
\begin{tabular}{lcccc}
\toprule \textbf{Corpus} & \textbf{\#Train} & \textbf{\#Dev} & \textbf{\#Test} & \textbf{Metrics}   \\ \midrule
\multicolumn{5}{l}{\emph{Single-Sentence Tasks}} \\
CoLA & 8.5k & 1k & 1k & Matthews Corr \\
SST-2 & 67k & 872 & 1.8k & Accuracy \\ 
\midrule
\multicolumn{5}{l}{\emph{Similarity and Paraphrase Tasks}} \\
QQP & 364k & 40k & 391k & Accuracy/F1 \\ 
MRPC & 3.7k & 408 & 1.7k & Accuracy/F1\\ 
STS-B & 7k & 1.5k & 1.4k & Pearson/Spearman Corr \\ 
\midrule
\multicolumn{5}{l}{\emph{Inference Tasks}} \\
MNLI & 393k & 20k &20k & Accuracy\\
RTE &2.5k & 276 & 3k & Accuracy \\ 
QNLI & 105k & 5.5k & 5.5k & Accuracy\\
WNLI & 634 & 71 & 146 & Accuracy\\ 
\bottomrule
\end{tabular}
\vskip -0.1in
\label{tbl:glue:datasets}
\end{table}

\begin{table}[t]
\caption{Dataset statistics and metrics of SQuAD 2.0.
}
\vskip 0.1in
\centering
\small
\begin{tabular}{cccc}
\toprule
\textbf{\#Train} & \textbf{\#Dev} & \textbf{\#Test} & \textbf{Metrics}   \\ \midrule
130,319 & 11,873 & 8,862 & Exact Match/F1 \\
\bottomrule
\end{tabular}
\vskip -0.1in
\label{tbl:squad2}
\end{table}

\section{SQuAD 2.0}

\paragraph{Question Generation} For the question generation task, we set batch size to $32$, and total length to $512$. The maximum output length is $48$. The learning rates are 3e-5 and 8e-5 for the $12$-layer and $6$-layer models, respectively. They are both fine-tuned for $25$ epochs. We also use label smoothing~\cite{label:smoothing} with rate of $0.1$. During decoding, we use beam search with beam size of $5$. The length penalty~\cite{gnmt} is $1.3$.

\paragraph{Abstractive Summarization} For the abstractive summarization task, we set batch size to $64$, and the rate of label smoothing to $0.1$. For the CNN/DailyMail dataset, the total length is $768$ and the maximum output length is $160$. The learning rates are 1e-4 and 1.5e-4 for the $12$-layer and $6$-layer models, respectively. They are both fine-tuned for $25$ epochs. During decoding, we set beam size to $5$, and the length penalty to $0.7$. For the XSum dataset, the total length is $512$ and the maximum output length is $48$. The learning rates are 1e-4 and 1.5e-4 for the $12$-layer and $6$-layer models, respectively. We fine-tune $30$ epochs for the $12$-layer model and $50$ epochs for the $6$-layer model. During decoding, we use beam search with beam size of $5$. The length penalty is set to $0.9$.

\paragraph{Cross-lingual Natural Language Inference} The maximum sequence length is $128$ for XNLI. We fine-tune $5$ epochs using $128$ as the batch size, choose learning rates from \{3e-5, 4e-5, 5e-5, 6e-5\}.

\paragraph{Cross-lingual Question Answering} For MLQA, the maximum sequence length is $512$ and a sliding window of size $128$ if the lengths are longer than $512$. We fine-tune $3$ epochs using $32$ as the batch size. The learning rates are chosen from \{3e-5, 4e-5, 5e-5, 6e-5\}.

\end{document}

%% file: settings.tex
\usepackage{multirow}
\usepackage{amsmath}
\usepackage{capt-of}
\usepackage{tabularx}
\usepackage{epsfig}
\usepackage{amssymb}
\usepackage{amsfonts}
\usepackage{scalerel}
\usepackage[inline]{enumitem}
\usepackage{listings}
\usepackage{varwidth}
\usepackage[export]{adjustbox}
\usepackage{tikz}
\usetikzlibrary{tikzmark}
\usepackage{todonotes}
\usepackage{cleveref}

\usepackage{stmaryrd}
\usepackage{bbm}

\newcommand{\tabincell}[2]{\begin{tabular}{@{}#1@{}}#2\end{tabular}}
\newcommand{\tx}[1]{``\textit{#1}''}
\newcommand{\sptk}[1]{\texttt{[#1]}}

\definecolor{mydarkblue}{rgb}{0,0.08,0.45}
\definecolor{deepblue}{rgb}{0,0,0.5}
\definecolor{officeblue}{RGB}{0,102,204}
\definecolor{deepred}{rgb}{0.6,0,0}
\definecolor{deepgreen}{rgb}{0,0.5,0}
\definecolor{mybrickred}{RGB}{182,50,28}

\definecolor{fillcolor}{RGB}{216,217,252}

%% file: math_commands.tex
\usepackage{amsmath,amsfonts,bm}

\def\eqref#1{equation~\ref{#1}}

\def\1{\bm{1}}

\DeclareMathAlphabet{\mathsfit}{\encodingdefault}{\sfdefault}{m}{sl}
\SetMathAlphabet{\mathsfit}{bold}{\encodingdefault}{\sfdefault}{bx}{n}

\newcommand{\softmax}{\mathrm{softmax}}

